\documentclass[sigconf]{aamas} 

\usepackage{balance} 

\usepackage{array}
\usepackage{graphicx} 
\newcolumntype{P}[1]{>{\centering\arraybackslash}p{#1}}
\usepackage{makecell}
\usepackage{enumitem}
\usepackage{multicol}
\usepackage{multirow}

\title{Learning Rewards, Not Labels: Adversarial Inverse Reinforcement Learning for Machinery Fault Detection}

\subtitle{Extended Abstract}

\author{Dhiraj Neupane}
\affiliation{
  \institution{Deakin University}
  \city{Waurn Ponds, Victoria}
  \country{Australia}}
  \orcid{0000-0001-6548-311X}
\email{d.neupane@deakin.edu.au}

\author{Richard Dazeley}
\affiliation{
  \institution{Deakin University}
  \city{Waurn Ponds, Victoria}
  \country{Australia}}
  \orcid{0000-0002-6199-9685}
\email{richard.dazeley@deakin.edu.au}

\author{Mohamed Reda Bouadjenek}
\affiliation{
  \institution{Deakin University}
  \city{Waurn Ponds, Victoria}
  \country{Australia}}
  \orcid{0000-0003-1807-430X}
\email{reda.bouadjenek@deakin.edu.au}

\author{Sunil Aryal}
\affiliation{
  \institution{Deakin University}
  \city{Waurn Ponds, Victoria}
  \country{Australia}}
  \orcid{0000-0002-6639-6824}
\email{sunil.aryal@deakin.edu.au}

\begin{abstract}
Reinforcement learning (RL) offers significant promise for machinery fault detection (MFD). However, most existing RL-based MFD approaches do not fully exploit RL's sequential decision-making strengths, often treating MFD as a simple \textit{guessing game} (Contextual Bandits). To bridge this gap, we formulate MFD as an offline inverse reinforcement learning problem, where the agent learns the reward dynamics directly from healthy operational sequences, thereby bypassing the need for manual reward engineering and fault labels. Our framework employs \textit{Adversarial Inverse Reinforcement Learning} to train a discriminator that distinguishes between normal (expert) and policy-generated transitions. The discriminator’s learned reward serves as an anomaly score, indicating deviations from normal operating behaviour. When evaluated on three run-to-failure benchmark datasets (HUMS2023, IMS, and XJTU-SY), the model consistently assigns low anomaly scores to normal samples and high scores to faulty ones, enabling early and robust fault detection. By aligning RL’s sequential reasoning with MFD’s temporal structure, this work opens a path toward RL-based diagnostics in data-driven industrial settings.
\end{abstract}

\keywords{Adversarial Learning; Anomaly Detection; Machinery Fault Detection; Reinforcement Learning; Prediction}

\newcommand{\BibTeX}{\rm B\kern-.05em{\sc i\kern-.025em b}\kern-.08em\TeX}

\renewcommand\footnotetextcopyrightpermission[1]{} 

\begin{document}


\pagestyle{fancy}
\fancyhead{}


\maketitle 
\enlargethispage{\baselineskip}
\textbf{* Accepted for publication at the 25th International Conference on Autonomous Agents and Multiagent Systems (AAMAS 2026). DOI - https://doi.org/10.65109/AXYX4522}
\vspace{1em}
\section{Introduction}
Machinery fault detection (MFD) is essential for maintaining industrial reliability, yet acquiring extensive labelled fault data remains a major bottleneck. While supervised learning dominates the field ($\approx$81\% of studies \cite{neupane2024comparative}), it suffers significantly from the scarcity of fault labels in real-world settings. Reinforcement learning (RL) has emerged as a promising alternative, aiming to model the sequential nature of degradation. However, most existing RL-based MFD approaches \cite{ding2019intelligent, qian2022development, li2025convolutional} reduce the problem to a static \textit{guessing game} or \textit{contextual-bandit} (CB) task. In these setups, agents treat sensor samples as independent states, issue one-shot classification actions, and ignore the discount factor ($\gamma=0$), thereby discarding the temporal structure inherent in fault progression \cite{neupane2024machineryDS}.

This simplification violates RL’s core premise of sequential decision-making. To bridge this gap, we propose formulating MFD as an offline \textit{Inverse Reinforcement Learning} (IRL) problem. Unlike standard RL, which requires a manually specified reward function, a significant challenge in complex machinery, IRL learns the reward function directly from expert demonstrations \cite{ng2000algorithms}.

We introduce an \textit{Adversarial Inverse Reinforcement Learning} (AIRL) \cite{fu2018airl} framework that learns the reward dynamics of \textit{healthy} machine operation. By treating normal operational sequences as ``expert" trajectories, our discriminator learns to distinguish between healthy transitions and generated anomalies. The learned reward function acts as an interpretable anomaly score: high rewards indicate alignment with healthy dynamics, while low rewards signal deviations. To the best of our knowledge, this is the first application of AIRL to MFD. Extensive experiments on three run-to-failure benchmarks (HUMS2023 \cite{wang2023helicopter}, IMS \cite{qiu2006wavelet}, XJTU-SY \cite{wang2020xjtusy}) demonstrate that our framework enables early and robust fault detection without requiring fault labels, outperforming traditional one-class and reconstruction-based baselines.

\section{Methodology}
We formulate MFD as an \textit{offline IRL problem } where the goal is to recover a reward function that rationalizes the behavior of a healthy (normal) machinery (the \textit{expert}).

\subsection{State Transition Construction}
Since industrial fault datasets lack recorded control inputs, we adopt a \textit{State-Only Imitation Learning (SOIL)} formulation \cite{torabi2019adversarial}. We segment the normalized vibration signals into fixed-length windows. Because explicit control actions are absent, to apply Inverse RL in this action-free setting, we define the state $s_t$ as the current window and treat the \textit{system's natural temporal evolution} to the next window as a ``proxy action'' ($a_t = x_{t+1}$). This formulation allows the AIRL discriminator to evaluate the plausibility of the transition dynamics $(s_t \to s_{t+1})$ by scoring them against the healthy expert distribution.

\subsection{Adversarial Reward Learning}
We employ AIRL \cite{fu2018airl}, which frames reward learning as a GAN-like optimization with two components: a \textit{Generator ($\pi$)} trained to mimic expert dynamics, and a \textit{Discriminator ($D$)}. 

The discriminator $D(s,a,s')$ estimates the probability that a transition is from the healthy expert distribution rather than the generator. Crucially, to recover a meaningful signal, AIRL structures the discriminator as:
\begin{equation}
    D(s,a,s') = \sigma\left(r_{\theta}(s,a) + \gamma V_{\phi}(s') - V_{\phi}(s) - \log \pi(a|s)\right)
\end{equation}
where $\sigma$ is the sigmoid function. This structural constraint forces the learned term $r_{\theta}(s,a)$ to act as a robust reward function (or health score), disentangled from the system dynamics.

\subsection{Anomaly Scoring}
Once trained, the discriminator $D$ estimates the probability that a transition belongs to the healthy manifold. High values indicate alignment with expert dynamics, while low values signal ``surprising'' deviations. We quantify this by defining the anomaly score for a trajectory $\tau$ as the inverted average discriminator confidence:
\begin{equation}
    \text{Score}(\tau) = 1 - \frac{1}{T} \sum_{t=0}^{T} D(s_t, a_t, s_{t+1})
\end{equation}
Fault onset is then identified by thresholding this score using dynamic methods (e.g., Otsu's method \cite{otsu1979threshold}, K-means \cite{macqueen1967kmeans}) and standard statistical rules.

\section{Experiments}
\subsection{Experimental Setup}

\begin{figure}
\centering
\includegraphics[width=0.48\textwidth]{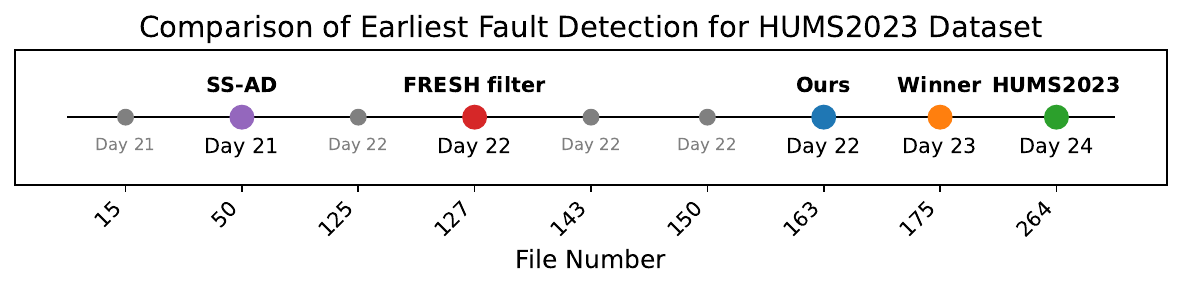}
\caption{Earliest Detection Onset Comparison (HUMS2023)}
\label{fig:hums_comparison}
\end{figure}

We evaluated the framework on three run-to-failure benchmark datasets: \textit{HUMS2023} (helicopter gearbox fatigue), \textit{IMS}, and \textit{XJTU-SY}.
For the primary HUMS2023 dataset, we selected the \textit{Ring-Front 2 (RF2)} accelerometer, as it best captures the fault-sensitive gear-meshing dynamics. The models were trained exclusively on healthy data (Days 17--20) and tested on the degradation phase (Days 21--27).
We compared AIRL against standard baselines (Isolation Forest (IF) \cite{liu2008isolation}, one-class support vector machines (OCSVM) \cite{scholkopf2001estimating}), reconstruction models (Autoencoder (AE), Variational Autoencoder (VAE)), and temporal reconstruction methods (LSTM-AE, LSTM-VAE). Additionally, we benchmarked against recent state-of-the-art methods including \textit{SS-AD} \cite{neupane2024comparative} and the \textit{FRESH-filter} \cite{vaerenberg2025detecting}, as well as a \textit{Contextual Bandit (CTQN)} \cite{li2025convolutional} baseline representing current RL-based MFD approaches.

\begin{table}[h]
\centering
\small
\caption{Earliest fault detection on HUMS2023.}
\label{tab:hums_detection}
\resizebox{\columnwidth}{!}{%
\begin{tabular}{lc|lc}
\toprule
\textbf{Model} & \textbf{Detection} & \textbf{Model} & \textbf{Detection} \\
\midrule
IF & Day 21 (\#9) & {VAE} & {Day 21 (\#37)}  \\
OCSVM & Day 21 (\#37) &  {LSTM-VAE} & {Day 22 (\#131)} \\
AE & Day 21 (\#1) & SS-AD \cite{neupane2024comparative} & Day 21 (\#50) \\
LSTM-AE     & Day 22 (\#131) & FRESH filter \cite{vaerenberg2025detecting} & Day 22 (\#127) \\
\textit{{CTQN (CB)}} & {{\textit{No Fault}}} & \textit{CW} \cite{peeters2024fatigue} & \textit{Day 23 (\#175)} \\
\midrule
\textit{Committee GT} & \textit{Day 24 (\#264)} & \textbf{AIRL (Ours)} & \textbf{Day 22 (\#163)} \\
\bottomrule
\multicolumn{4}{l}{\footnotesize \textit{GT: Ground Truth; CW: Challenge Winner}} \\
\end{tabular}%
}
\end{table}

\subsection{Results}
The primary evaluation metric was the \textit{earliest valid detection} of fault onset. Due to space constraints, we detail the detection results for the primary HUMS2023 dataset in Table~\ref{tab:hums_detection}; however, consistent performance trends were observed across the IMS and XJTU-SY benchmarks.

{Our \textit{AIRL framework} identified the fault onset at \textit{Day 22 (File \#163)}. As visualized in Figure~\ref{fig:hums_comparison}, this detection falls squarely between the FRESH filter (File \#127) and the official \textit{Challenge Winner} (Day 23, File \#175) \cite{peeters2024fatigue}. Crucially, our detection precedes the conservative ground truth established by the HUMS committee (Day 24, File \#264), demonstrating that AIRL provides a valuable early warning window without the premature false positives observed in other methods. Beyond timely detection, AIRL demonstrated superior \textit{post-Detection Consistency (PDC)}, maintaining a steady anomaly rate ($\approx$65\%) after fault onset. This stability was mirrored in the IMS and XJTU-SY experiments, confirming the robustness of the sequential reward formulation.}

{In comparison, as shown in Table~\ref{tab:hums_detection}, standard baselines (IF, OCSVM, AE) flagged anomalies prematurely. While sequential models like LSTM-AE and LSTM-VAE improved precision (Day 22), they still triggered earlier than our method. Crucially, the {Contextual Bandit} (CTQN) baseline \textit{failed entirely}, classifying the whole test set as normal. This confirms that without modeling state transitions ($\gamma=0$), the agent cannot perceive the gradual accumulation of fatigue damage.}

\section{Conclusion}

This work introduces the first application of Adversarial Inverse Reinforcement Learning to machinery fault detection. Unlike existing ``RL-based'' methods that treat fault diagnosis as a static \textit{contextual-bandit} problem, our framework respects the sequential nature of machine degradation. By recovering a reward function from healthy data, AIRL provides a robust, interpretable anomaly score that detects faults early and reliably.
Our results on HUMS2023, IMS, and XJTU-SY demonstrate that learning the \textit{dynamics} of health is superior to merely classifying isolated observations. Future work will extend this framework to multi-sensor fusion and incorporate uncertainty-aware thresholding to further reduce false alarms in variable operating conditions.

\balance
\bibliographystyle{unsrt}
\bibliography{sample}



\end{document}